\documentclass[10pt,letterpaper]{article}

\usepackage{cogsci}

\cogscifinalcopy 

\usepackage{pslatex}
\usepackage{apacite}
\usepackage{float} 
\usepackage{graphicx}
\usepackage{multirow}
\usepackage{booktabs}

\usepackage{amsfonts}       
\usepackage{amsmath}        
\usepackage{algorithm}      
\usepackage{algpseudocode}  
\usepackage{listings}       
\usepackage{enumitem}       
\usepackage{xcolor}         
\usepackage{float}          
\usepackage{subcaption}     

\title{Modeling the Mistakes of Boundedly Rational Agents \\ Within a Bayesian Theory of Mind}

\author{
{\large \bf
    Arwa Alanqary\textsuperscript{\textasteriskcentered},
    Gloria Z. Lin\textsuperscript{\textasteriskcentered},
    Joie Le\textsuperscript{\textasteriskcentered},
    Tan Zhi-Xuan\textsuperscript{\textasteriskcentered\textdagger}} \\
{\large \bf
    Vikash K. Mansinghka,
    Joshua B. Tenenbaum} \\
  Department of Brain and Cognitive Sciences, MIT \\
  \texttt{\{alanqary,gzlin,joiele,xuan,vkm,jbt\}@mit.edu} \\ \\
{\small
    \textsuperscript{*}Equal Contribution \quad
    \textsuperscript{$\dagger$}Corresponding Author}
}

\begin{document}

\maketitle

\begin{abstract}
When inferring the goals that others are trying to achieve, people intuitively understand that others might make mistakes along the way. This is crucial for activities such as teaching, offering assistance, and deciding between blame or forgiveness. However, Bayesian models of theory of mind have generally not accounted for these mistakes, instead modeling agents as mostly optimal in achieving their goals. As a result, they are unable to explain phenomena like locking oneself out of one's house, or losing a game of chess. Here, we extend the Bayesian Theory of Mind framework to model \emph{boundedly} rational agents who may have mistaken goals, plans, and actions. We formalize this by modeling agents as probabilistic programs, where goals may be confused with semantically similar states, plans may be misguided due to resource-bounded planning, and actions may be unintended due to execution errors. We present experiments eliciting human goal inferences in two domains: (i) a gridworld puzzle with gems locked behind doors, and (ii) a block-stacking domain. Our model better explains human inferences than alternatives, while generalizing across domains. These findings indicate the importance of modeling others as bounded agents, in order to account for the full richness of human intuitive psychology.

\textbf{Keywords:} Bayesian Theory of Mind, Action Understanding, Planning, Bounded Rationality, Social Cognition

\end{abstract}

\section{Introduction}
A key aspect of human intuitive psychology is our understanding that other agents are fallible: they may possess false beliefs \cite{wimmer1983beliefs}, lack knowledge \cite{phillips2019we}, fail to plan ahead \cite{callaway2018resource}, or act unintentionally \cite{schult1997explaining}. This capacity is crucial to social life, allowing us to teach others \cite{wellman2004theory}, or forgive harms that we take as unintended \cite{young2007neural,cushman2008crime}. Remarkably, even 18-month old infants seem to account for such errors when inferring the goals of others, enabling them to offer assistance \cite{warneken2006altruistic}.

What are these errors, and how do we understand them in a way that allows us to infer the goals and intentions of others? In the Bayesian Theory of Mind (BToM) framework, goal inference is explained as a form of \emph{inverse planning}, where observers infer a posterior distribution over goals by modeling agents as rational planners \cite{baker2009action}. However, while prior BToM models successfully explain how humans can infer others' goals, desires, and intentions \cite{jara2016naive,liu2017ten}, and have even explained how we might infer mistaken beliefs \cite{baker2017rational,evans2016learning}, little attention has been paid to mistaken goals, plans, or actions. With a few exceptions \cite{evans2015learning,kryven2016outcome}, most BToM models only account for mistakes through Boltzmann-distributed action noise \cite{ziebart2008maximum}. This fails to capture higher-level mistakes, and has been challenged as a model of sequential decision making \cite{otter2008sequential,bobu2020less}.

In this paper, we build upon a recently proposed model of agents as boundedly rational planners \cite{zhi2020online}. Unlike earlier BToM agents which plan via exhaustive computation of expected value over the entire state space \cite{ramachandran2007bayesian,baker2009action}, these agents do not always plan optimally, but, like ourselves, only plan several steps ahead before executing that partial plan and replanning. This is resource rational in many cases \cite{callaway2018resource,bratman1988plans}, but can also lead to failure: you might lock yourself out of the house, because you neglect to bring your keys. We extend this model with goal mistakes, due to confusion of goals with semantically similar specifications, and action mistakes, due to occasional execution of unplanned actions. Our model thus accounts for sub-optimality at three distinct levels of human decision-making.

We hypothesize that human goal inferences given sub-optimal action sequences are better explained by Bayesian inference in this model that accounts for goal, plan and action mistakes than previously introduced BToM approaches. We test this hypothesis by eliciting human inferences in two environments: (i) a gridworld puzzle domain called Doors, Keys \& Gems, and (ii) a variant of the well-known Blocks World domain where an agent spells words with lettered blocks. In these experiments, participants observe an agent as it progresses towards its goal, but makes mistakes along the way. At selected points, they are asked to judge which goals are the most likely. Observed sequences are designed to exhibit mistakes at the levels of goals, plans, and actions. By comparing human judgements against inferences for each computational model, we evaluate the fidelity of these models to our intuitive theory of mind.

\section{Computational Model}

To account for mistakes at multiple levels of decision making, we model agents and their environments as generative processes of the following form:
\begin{alignat}{2}
\textit{Goal prior:}& \qquad g_0 \sim P(g_0) \label{eq:goal-prior} \\
\textit{Goal transition:}& \qquad  g_t \sim P(g_t|g_{t-1}, g_0) \label{eq:goal-transition} \\
\textit{Plan update:}& \qquad p_t \sim P(p_t | s_{t-1}, p_{t-1}, g_{t-1}) \label{eq:plan-update} \\
\textit{Action selection:}& \qquad a_t \sim P(a_t | s_t, p_t) \label{eq:action-selection} \\
\textit{State transition:} & \qquad s_t \sim P(s_t | s_{t-1}, a_t) \label{eq:state-transition} \\
\textit{Observation noise:}& \qquad o_t \sim P(o_t|s_t) \label{eq:observation-noise}
\end{alignat}
where $g_0$ is the agent's original intended goal, and $g_t$, $p_t$, $a_t$, $s_t$ are the agent’s current (potentially corrupted) goal, the internal state of the agent’s plan, the agent’s action, and the environment’s state at time $t$ respectively.  These generative processes are specified as probabilistic programs (Figure \ref{fig:agent-subroutines}), and their corresponding mistakes are described below.

\begin{figure}[t]
\centering
\includegraphics[width=\columnwidth]{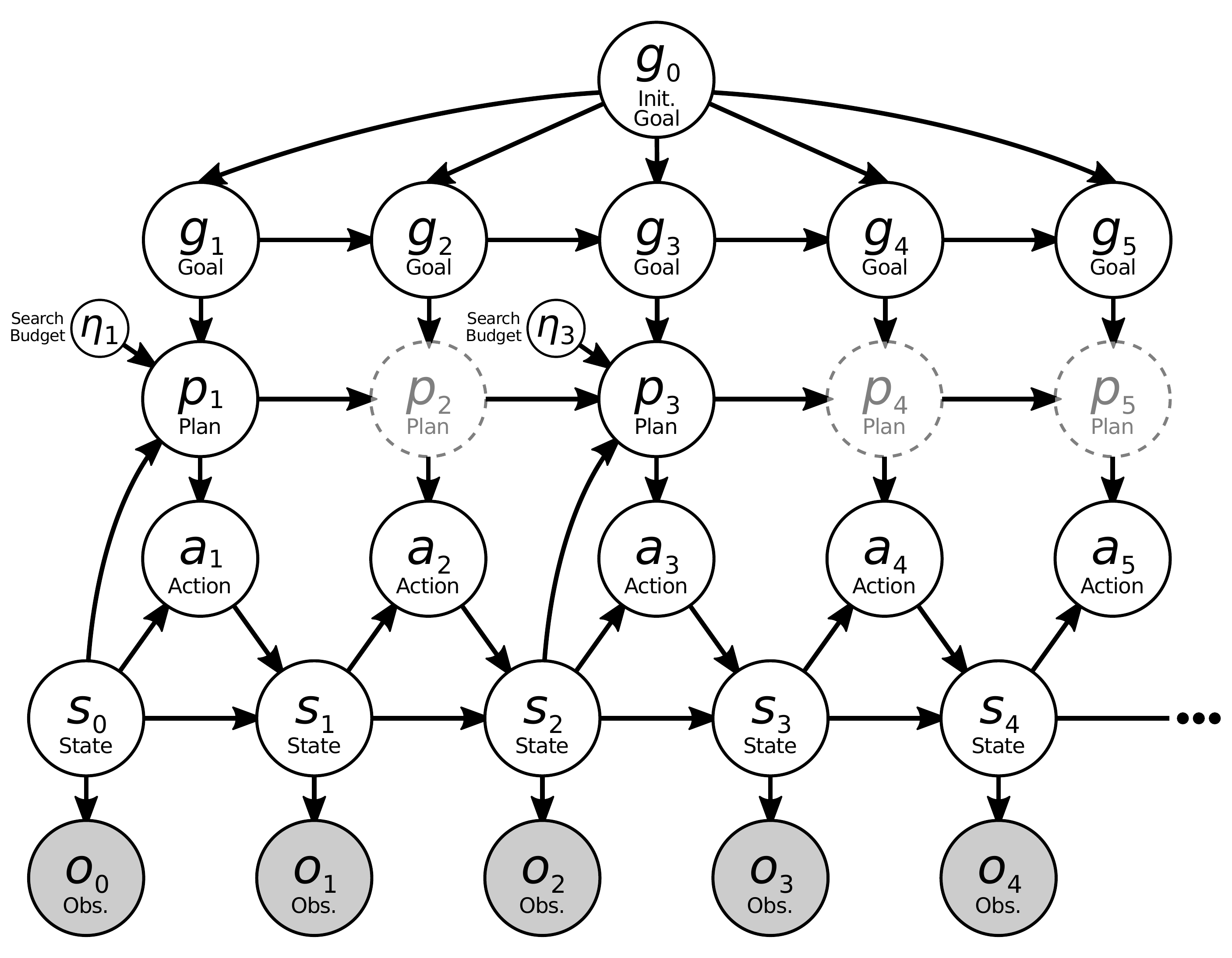}
\caption{A single realization of our boundedly rational agent model. At $t$=$1$, the agent samples a search budget $\eta_1$ and searches for a plan $p_1$ that is two actions long. At $t$=$2$, no additional planning needs to be done, so $p_2$ is copied from $p_1$, as denoted by the dashed lines. The agent then replans at $t$=$3$, sampling a new search budget $\eta_3$ and an extended plan $p_3$ with three more actions.}
\label{fig:graphical-model}
\end{figure}

\begin{figure}[t]
\centering
\algrenewcommand\algorithmicprocedure{\textbf{model}}
\begin{algorithmic}[1]
\footnotesize
\Procedure{goal-transition}{$t$, $g_{t-1}$, $g_0$}
\State \textbf{parameters}: $\epsilon_g$ (goal noise)
\If{$\textsc{bernoulli}(\epsilon_g) = \textsc{true}$}
    \State \Return \textbf{if} $g_0 = g_{t-1}$ \textbf{then} $\textsc{corrupt}(g_0)$
    \textbf{else} $g_0$ \textbf{end if}
\Else
    \State \Return $g_{t-1}$
\EndIf
\EndProcedure
\end{algorithmic}
\footnotesize{\textbf{(i)} Samples goals from $P(g_t | g_{t-1}, g_0)$}

\vspace{3pt}
\begin{algorithmic}[1]
\footnotesize
\Procedure{plan-update}{$t$, $s_t$, $p_{t-1}$, $g$}
\State \textbf{parameters}: \textsc{planner}, $r, q, \gamma, h$
\If{$t > \textsc{length}(p_{t-1})$ \textbf{or} $s_t \notin p_{t-1}[t]$}
    \State $\eta \sim \textsc{negative-binomial}(r, q)$
    \State $\tilde{p}_t \sim \textsc{planner}(s_t, g, h, \gamma, \eta)$
    \State $p_t \gets \textsc{append}(p_{t-1}, \tilde{p}_t)$
\Else
    \State $p_t \gets p_{t-1}$
\EndIf
\State \Return $p_t$
\EndProcedure
\end{algorithmic}
\footnotesize{\textbf{(ii)} Samples plans from $P(p_t | s_t, p_{t-1}, g)$}

\vspace{3pt}
\begin{algorithmic}[1]
\footnotesize
\Procedure{action-selection}{$t$, $s_t$, $p_t$}
\State \textbf{parameters}: $\epsilon_a$ (action noise)
\If{$\textsc{bernoulli}(\epsilon_a) = \textsc{true}$}
    \State \Return $\textsc{uniform}(\textsc{actions}(s_t) \setminus p_t[t][s_t])$
\Else
    \State \Return $p_t[t][s_t]$
\EndIf
\EndProcedure
\end{algorithmic}
\footnotesize{\textbf{(iii)} Samples actions from $P(a_t | s_t, p_t)$}
\vspace{1pt}
\caption{Generative subroutines specifying (i) goal transition noise, (ii) plan updates, and  (iii) action selection.}
\label{fig:agent-subroutines}
\end{figure}

\subsection{Mistaken goals via temporary goal confusion}

Consistent with previous BToM literature \cite{baker2009action}, we restrict our scope to inference over a fixed set of context-relevant goals with associated prior probabilities (Eq. \ref{eq:goal-prior}), leaving aside how people come up with these contextual hypotheses. However, within a fixed set of goals, complex goals can still get confused. For example, when stacking blocks to spell a word, one might have in mind a misspelling: Is it ``firey'' or ``fiery''? To account for these mistakes, we introduce goal transition noise (Eq. \ref{eq:goal-transition}), specified by the procedure in Figure \ref{fig:agent-subroutines}(i). At each step $t$ with probability $\epsilon_g$, the original goal $g_0$ may be corrupted to produce a similar \emph{temporary} goal $g_t$, or the corrupted goal corrected to the original one (Line 4). Goal noise can be domain-specific, e.g. in Blocks World, \textsc{corrupt} might correspond to a random permutation.

\subsection{Mistaken plans via resource-bounded planning}

As described in the introduction, we model agents that interleave resource-bounded planning with plan execution. This can cause mistakes due to failure to plan ahead. At each step $t$, agents already have a previous partial plan $p_{t-1}$, and construct an updated plan $p_t$  (Eq. \ref{eq:plan-update}).

This plan update procedure is shown in Figure \ref{fig:agent-subroutines}(ii), and it captures boundedly rational planning in the following ways: First, the agent does not update it previous plan (i.e., a sequence of intended actions) if it already extends to the current step $t$ (Line 8). Second, if the plan \emph{does} need to be extended, the agent only spends a limited budget $\eta$ to construct a new partial plan $\tilde p_t$. For example, if the agent plans via forward search, $\eta$ is the number of steps the agent thinks ahead. We sample $\eta$ from a negative binomial distribution (Line 4), encoding the assumption that neither very short nor very long plans are likely.
Third, we assume that the planning procedure itself, \textsc{planner}, is noisy (Line 5). While in principle, any planning algorithm could be used, we adopt a probabilistic version of A* search, capturing the intuition that humans often plan by thinking a few steps ahead, guided by a heuristic $h(s, g)$ that evaluates how promising a state $s$ is relative to the goal $g$. In regular A* search, only the most promising state $s$ is expanded at each iteration. However, since humans may not rank states perfectly, we instead sample $s$ from the Boltzmann distribution:
    $$P_\text{expand} (s) \propto \exp(-f(s,g)/\gamma)$$
where higher $\gamma$ increases the randomness of search, $c(s)$ is the cost of reaching $s$ from the initial state, and $f(s,g) = c(s) + h(s, g)$ is the estimated total cost of reaching the goal $g$ by passing through $s$. The search algorithm terminates when either the goal state $g$ is reached or the $\eta$th state is expanded (i.e. the plan budget is exhausted), at which point a partial plan $\tilde p_t$ to last-expanded state $s$ is returned.

\subsection{Mistaken actions via execution errors}

Humans may not always execute plans as intended. Instead, due to carelessness or lack of motor control, we may sometimes commit execution errors, e.g., dropping a block by accident, or walking a step more than intended. Hence, we model action selection (Eq. \ref{eq:action-selection}, Figure \ref{fig:agent-subroutines}(iii)) as a process where the agent usually executes their intended action given the current plan $p_t$ and state $s_t$ (Line 6), but with probability $\epsilon_a$ randomly executes another possible actions in state $s_t$ (Line 4).



\subsection{Bayesian goal inference}

We model observers as performing Bayesian inference over an agent's intended goal $g_0$ given observations $o_{1:t}$. We assume that the agent and observer have a shared model of the environment, specified in the Planning Domain Definition Language \cite{mcdermott1998pddl}. Observations are provided as symbolic predicates, and to model observation noise, Boolean predicates are corrupted with probability $\epsilon_o$, while numeric predicates have Gaussian noise added with variance $\sigma_o^2$. Given the complexity of our model, exactly computing the goal posterior $P(g_0|o_{1:t})$ is intractable. Thus, we use Sequential Inverse Plan Search (SIPS), the sequential Monte Carlo algorithm developed in \citeA{zhi2020online}, to approximate $P(g|o_{1:t})$. We refer readers to that work for technical details.

As baselines for comparison, we compute goal inferences using lesioned agent models: \textbf{G-Lesioned}, where goal mistakes are absent, \textbf{P-Lesioned}, where resource bounded planning is absent, or \textbf{A-Lesioned}, where action mistakes are absent. We also compare with the \textbf{Boltzmann agent model} used in earlier BToM approaches \cite{baker2009action} and Bayesian Inverse Reinforcement Learning \cite{ramachandran2007bayesian}. In this model, agents precompute the expected future reward $V(s)$ of every state $s$, and follow a Boltzmann policy, noisily selecting actions that tend to maximize reward:
\begin{equation}
    \pi(a|s) \propto \exp( \alpha [R(s,a,s') + V(s')])
\end{equation}
Here, $s'$ is the successor state, $R(s,a,s')$ is the reward from taking action $a$ from $s$ to $s'$, and higher $\alpha$ leads to lower noise. We set $R(s,a,s')=-1$ for all actions and treat goal states as terminal, leading agents to prefer shorter routes to goals. Because this model exhaustively plans (i.e. computes the expected reward $V(s)$) over the entire state space, it only accounts for low-level action mistakes. As such, we hypothesize that it will not explain mistaken goals or plans as well as our boundedly-rational model.


\section{Experiments}

To demonstrate the generality of our model, we conducted experiments in two domains: (i) a gridworld puzzle called Doors, Keys \& Gems, and (ii) a Blocks World variant called Block Words, where an agent spells words out of lettered blocks. These domains exhibit the compositional structure that humans encounter in daily life, making them tractable to plan in, but also complex enough for mistakes to arise.

In each domain, we elicited goal inferences from human observers as they watched a variety of optimal and suboptimal agent trajectories unfold. Given the complexity of our model, sub-optimal trajectories might admit multiple interpretations:  ``Was that a mistaken action, or a mistaken plan?'' As such, we designed trajectories to make some mistakes more likely than others. For example, if someone walked a step out of the house before turning to get their keys, an observer might take that as a mistaken action, but if they walked all the way to the bus stop before turning around, a mistaken plan would seem more likely. In the Block Words domain specifically --- which has a richer goal space --- we also designed trajectories with mistaken goals: agents would sometimes stack a misspelled word, before correcting themselves.

\begin{figure*}[ht]
    \centering
    \begin{subfigure}[b]{0.53\textwidth}
         \centering
         \includegraphics[width=\textwidth]{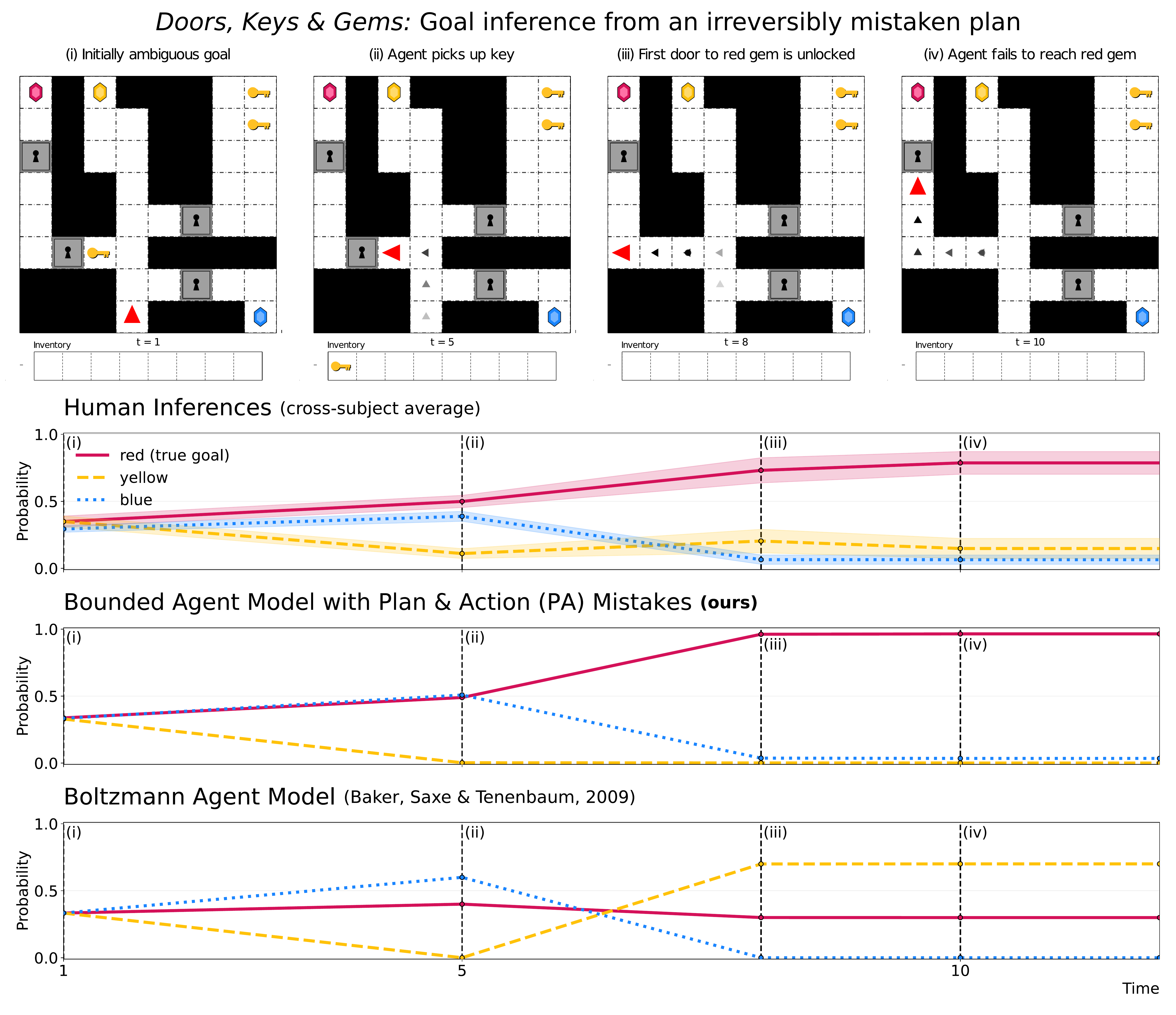}
         \caption{Goal inferences over time for a trajectory with an irreversible failure.}
         \label{fig:traj_dkg}
     \end{subfigure}
     \hspace{3pt}
     \begin{subfigure}[b]{0.45\textwidth}
         \centering
         \includegraphics[width=\textwidth]{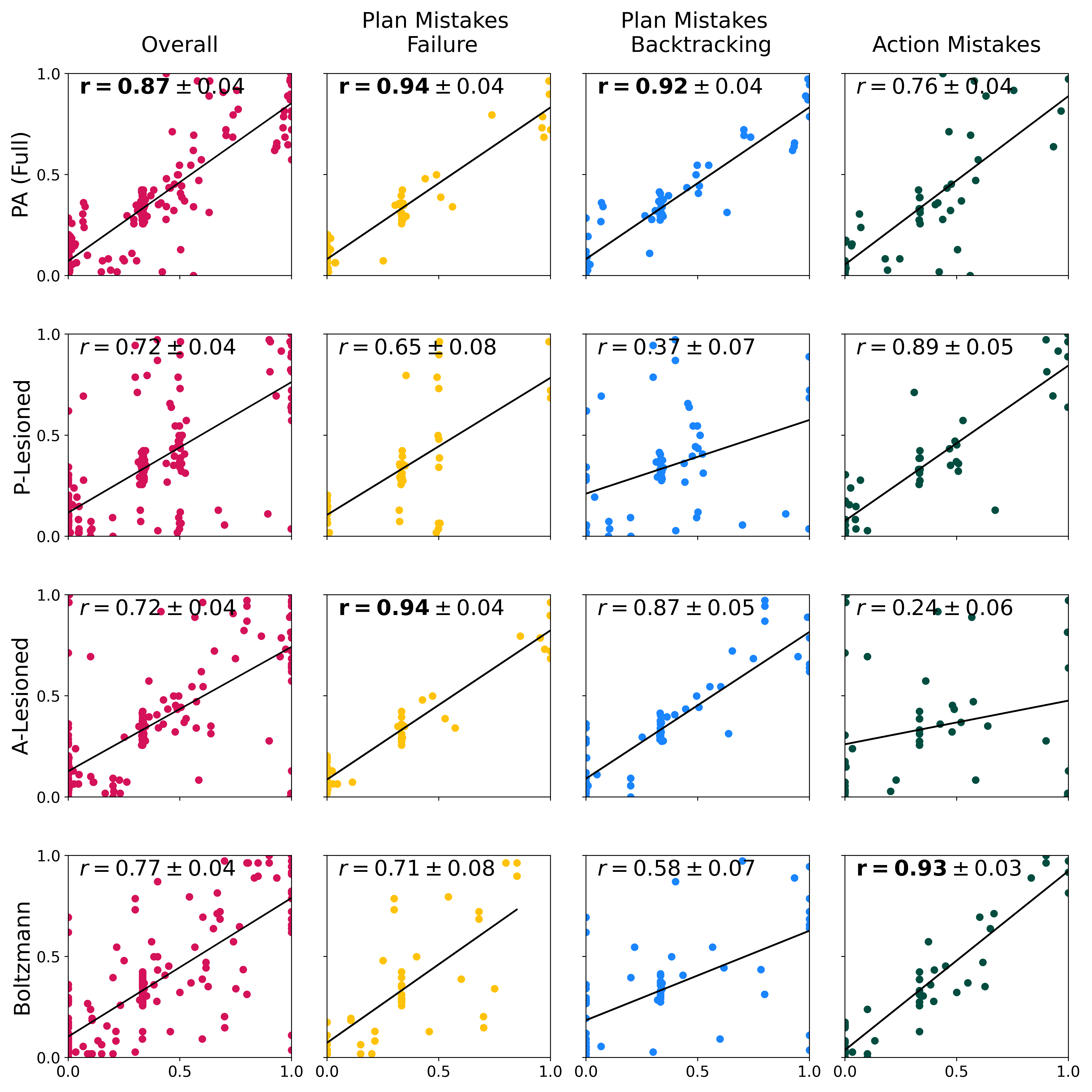}
         \caption{Human ($y$-axis) vs. model ($x$-axis) inferences.}
         \label{fig:scatter_dkg}
     \end{subfigure}
     \caption{Results for the Doors, Keys \& Gems domain. In \textbf{(a)}, we show goal inferences of humans (avg., $n$=18, w. standard error), our model (PA) and the Boltzmann agent model for an illustrative trajectory. In \textbf{(b)}, we compare human  vs. model inferences across models (full model, lesioned models for plan/action mistakes, Boltzmann model) and stimulus types.}
    \label{fig:doors-keys-gems-results}
\end{figure*}

\subsection{Experiment 1: Doors, Keys, \& Gems}

In this domain, an agent must navigate a maze in order to collect one of three colored gems, which may be locked behind doors (Figure \ref{fig:doors-keys-gems-results}(a)). Keys are required to unlock doors, and can only be used once, leading to the possibility of irreversible failure if the agent does not plan ahead. We assume that planning is guided by a maze distance heuristic that ignores the presence of doors, leading occasionally to myopic plans which neglect the need for keys. Because the goals in this domain are simple, we do not model mistaken goals.

\textbf{Stimuli.} We designed 16 agent trajectories as stimuli for human participants, organized into four subsets: (1) A control set of optimal trajectories.  (2) Trajectories with irreversibly mistaken plans, where the agent myopically uses up all obtainable keys, such that the goal gem is locked out of reach (Figure \ref{fig:doors-keys-gems-results}(a)). (3) Trajectories with mistaken actions, where the agent takes a few false steps, then corrects their behavior. (4) Trajectories with short-sighted plans, leading the agent to backtrack to obtain keys. Participants accessed a web interface which presented all 16 stimuli in random order. Stimuli were presented as animated videos which paused at selected judgement points. At each point, participants provided goal inferences by selecting the gem(s) they believed to be the agent's most likely goal. Participants could select multiple gems if more than one seemed equally likely, and these responses were converted to probability distributions.


\textbf{Participants.} We recruited 20 US participants (mean age 36.4, SD 10.4; 8 women, 12 men, 0 non-binary/other) via Amazon Mechanical Turk (AMT), restricting to those with a HIT approval rate of $99\%$ and above.  Participants went through a tutorial and answered four comprehension questions before viewing the stimuli. Participants also earned points proportional to the probability they gave the true goal (mean score 28, SD 10), and were paid \$1 per 10 points, incentivizing accurate guesses. Two participants were excluded from our analysis, either for failing two or more comprehension checks, or for guessing indiscriminately and failing to reach a threshold of 10 points.

\textbf{Analysis.} We fit the parameters of each model via grid search, maximizing the overall correlation (Pearson's $r$) between average human judgements and model inferences. Model inferences were computed by averaging 10 runs of the SIPS particle filter \cite{zhi2020online} with 300 particles per run. We then computed 95\% bootstrap confidence intervals for the $r$ values using 500 resamples of the human data. We also conducted a sensitivity analysis to validate whether any the models performed better just because of overfitting. Details and model parameters can be found in the Appendix.


\textbf{Results.}  Figure \ref{fig:doors-keys-gems-results} shows the results of our analysis. An illustrative example from the Doors, Keys, \& Gems domain is shown in Figure \ref{fig:doors-keys-gems-results}(a), where an agent locks their desired gem out of reach due to myopic planning. The panels show how the agent mistakenly uses up a key to unlock the first door to the red gem, instead of collecting the other two keys. They then approach the second door, and are stuck. Below these panels, we show average human goal inferences over time (with standard error ribbons), alongside the inferences for our boundedly-rational agent model and the baseline Boltzmann agent model. Humans are able to recognize the true goal as soon as the first door is unlocked ($t$=$8$) despite knowing that the agent won't be able to reach it. Our model, which accounts for plan and action (PA) mistakes, exhibits highly similar behavior to humans, placing high confidence in the red gem once the door is unlocked ($t$=$8$). For the Boltzmann agent model, however, the probability of the red gem decreases at $t$=$8$, and more weight goes to the yellow gem, since it becomes the only gem reachable without any keys.

We present correlation plots between average human and model inferences in Figure $\ref{fig:doors-keys-gems-results}$(b). The left column shows correlations across all 16 stimuli, with our full model (PA) best explaining human inferences ($r$=$0.87 \pm 0.04$). The next two columns show correlations for stimuli with mistaken plans, leading to failure and backtracking respectively. In both cases, our full model fit the human data best. The final column shows correlations only for stimuli with action mistakes, and here our model performed slightly worse than models that account only for action noise (P-Lesioned, Boltzmann). Notably, however, these models performed much more poorly on mistaken plans, mirroring how the A-Lesioned model performed much worse with mistaken actions. Sensitivity analysis (Figure \ref{fig:sensitivity}(a)) indicated that the high correlation for our full model was not just due to overfitting: Apart from a few parameter settings, our full model achieved consistently higher correlations than the Boltzmann or lesioned models.

\begin{figure*}[ht]
     \centering
     \begin{subfigure}[b]{0.53\textwidth}
         \centering
         \includegraphics[width=\textwidth]{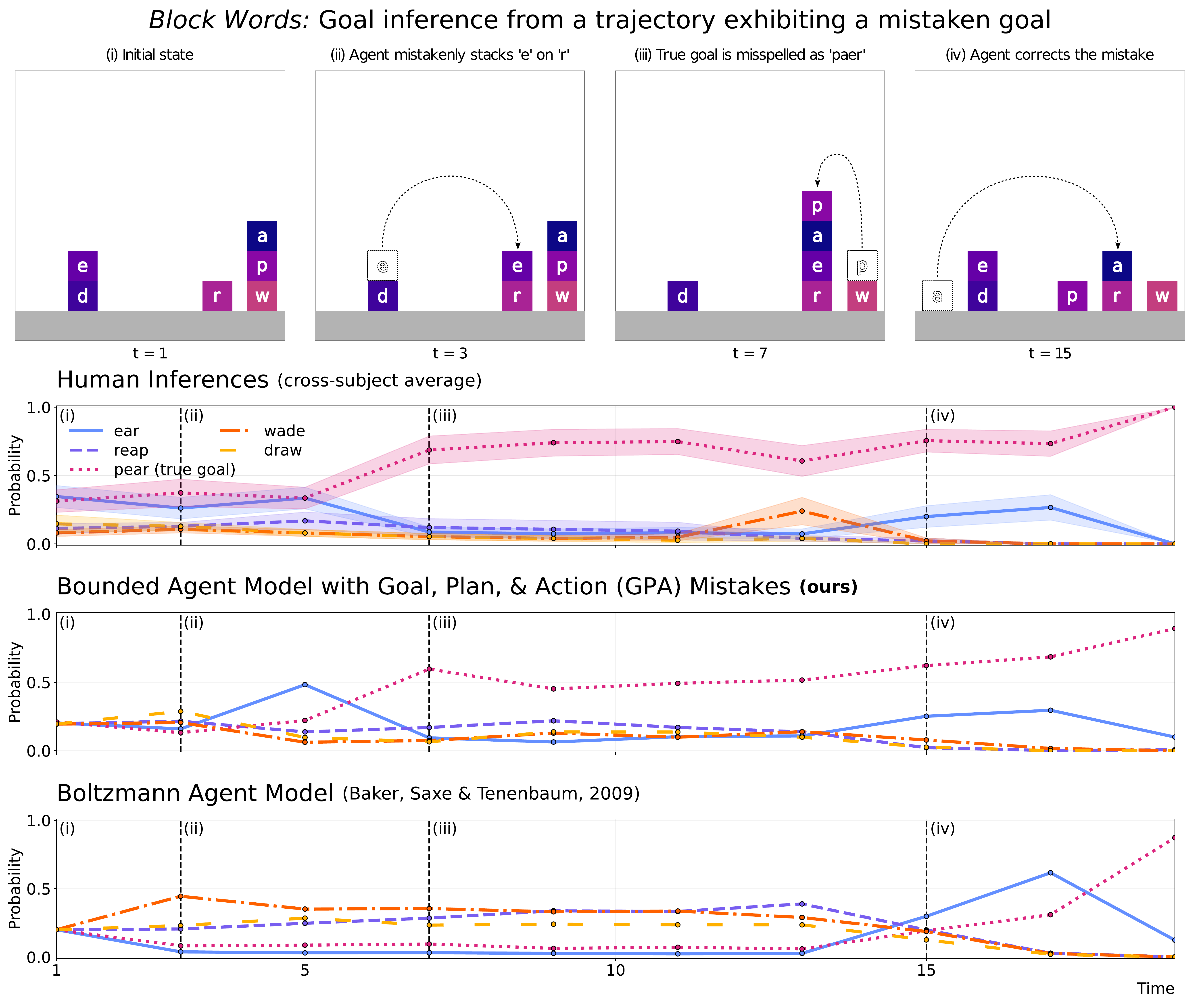}
         \caption{Goal inferences over time for a trajectory with a mistaken goal.}
         \label{fig:traj_bw}
     \end{subfigure}
     \hspace{5pt}
     \begin{subfigure}[b]{0.42\textwidth}
         \centering
         \includegraphics[width=\textwidth]{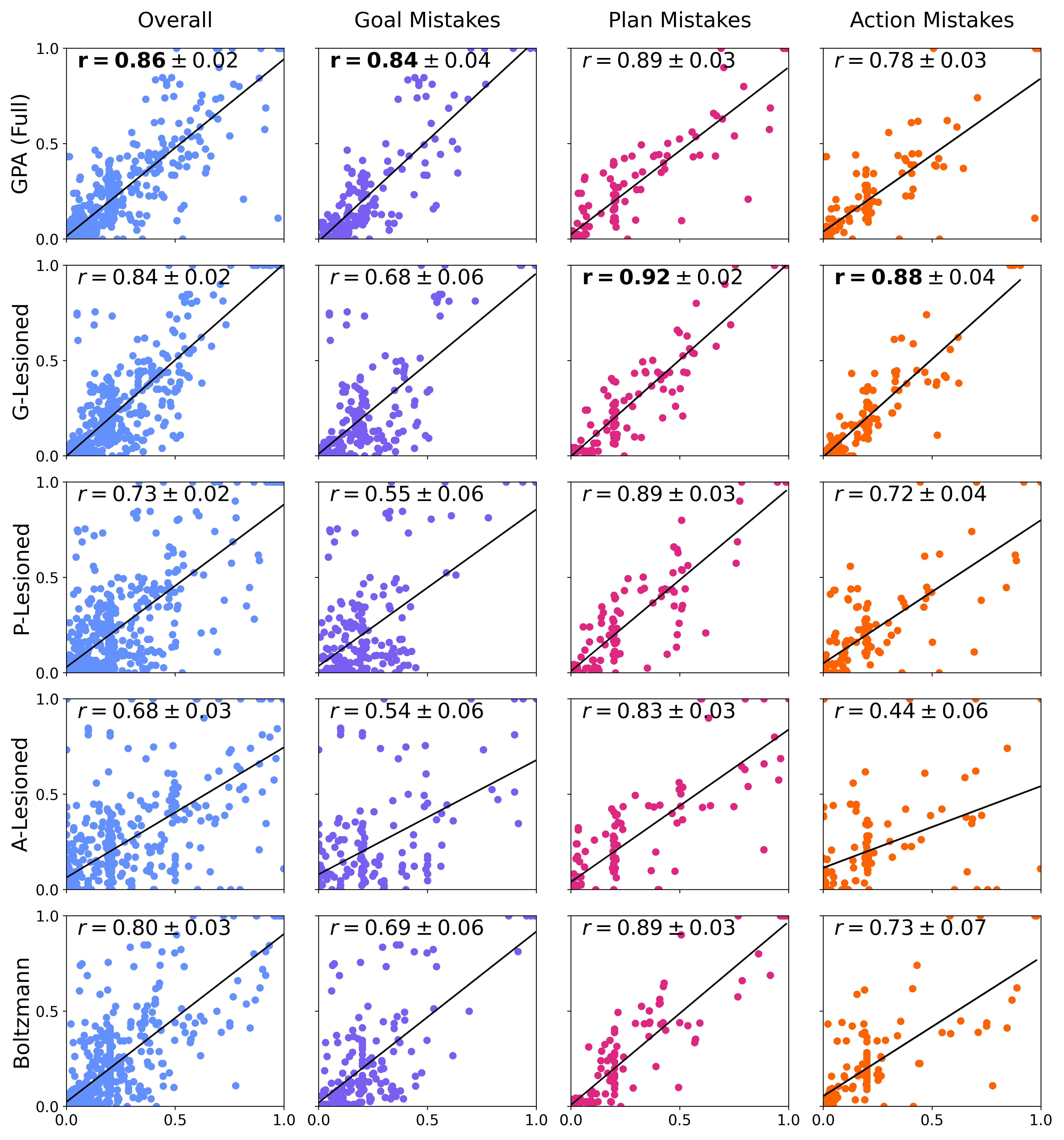}
         \caption{Human ($y$-axis) vs. model ($x$-axis) inferences.}
         \label{fig:scatter_bw}
     \end{subfigure}
     \caption{Results for the Block Words domain. In \textbf{(a)}, we show goal inferences of humans (avg., $n$=27, w. standard error), our model (GPA) and the Boltzmann agent model on an illustrative trajectory. In \textbf{(b)}, we compare human  vs. model inferences across models (full model, lesioned models for goal/plan/action mistakes, Boltzmann model) and stimulus types.}
    \label{fig:block-words-results}
\end{figure*}

\subsection{Experiment 2: Block Words}

In this domain adapted from \citeA{ramirez2010probabilistic}, blocks are labeled with letters, and an agent may pick up a block, put it down, or stack it on top of another. The agent's goal is to stack a block tower that spells (top-down) one out of five English words, which are provided to the observer in advance. We assume that planning is guided by a domain-general relaxed distance heuristic \cite{geffner2010heuristics}, leading agents to occasionally neglect how stacking one block on another will prevent the block underneath from being reached. Due to the complexity of goals in this domain, we account for temporary goal confusion, where the agent tries to spell a random permutation of the original word, for example, ``paer'' instead of ``pear'' (Figure \ref{fig:block-words-results}(a)).

\textbf{Stimuli.} As in Experiment 1, we designed 16 stimuli, organized into four subsets: (1) A control set of optimal trajectories. (2) Trajectories with mistaken goals (misspellings of the intended block tower). (3) Trajectories with mistaken plans, where the agent stacks a block on top of other block(s) it will later need. (4) Trajectories with mistaken actions, where the agent drops a block in an unintended location, or picks up a block adjacent to the one intended.  Participants accessed a web interface which presented 10 stimuli in random order, with at least two stimuli from each subset. Stimuli were presented as animations which paused every two actions (picking and placing a block). At these pauses, subjects selected which word(s) they believed to be the most likely goal. Participants could select multiple words if they seemed equally likely, and these responses were converted to probability distributions.

\textbf{Participants.} We recruited 32 US participants (mean age 40.8, SD 12.5; 13 women, 19 men, 0 non-binary/other) via AMT, restricting to those with a HIT approval rate of $99\%$ and above.
Participants went through a tutorial and answered five comprehension questions before proceeding to the stimuli. Following Experiment 1, we awarded points proportional to the probability they assigned to the true goal (mean score 37, SD 7). Five participants were excluded from our analysis, either for failing two or more comprehension checks, or for failing to reach a threshold of 20 points.

\textbf{Analysis.} Similar to Experiment 1, we fit the parameters of each model via grid search, computed 95\% bootstrap confidence intervals, and conducted a sensitivity analysis by examining how $r$ values changed across parameter sets. Model inferences were computed by averaging 10 runs of the SIPS particle filter with 500 particles per run (100 particles per potential goal, as in Experiment 1). Details and parameters can be found in the Appendix.

\begin{figure*}[ht]
    \centering
    \begin{subfigure}[b]{0.4\textwidth}
         \centering
         \includegraphics[width=\textwidth]{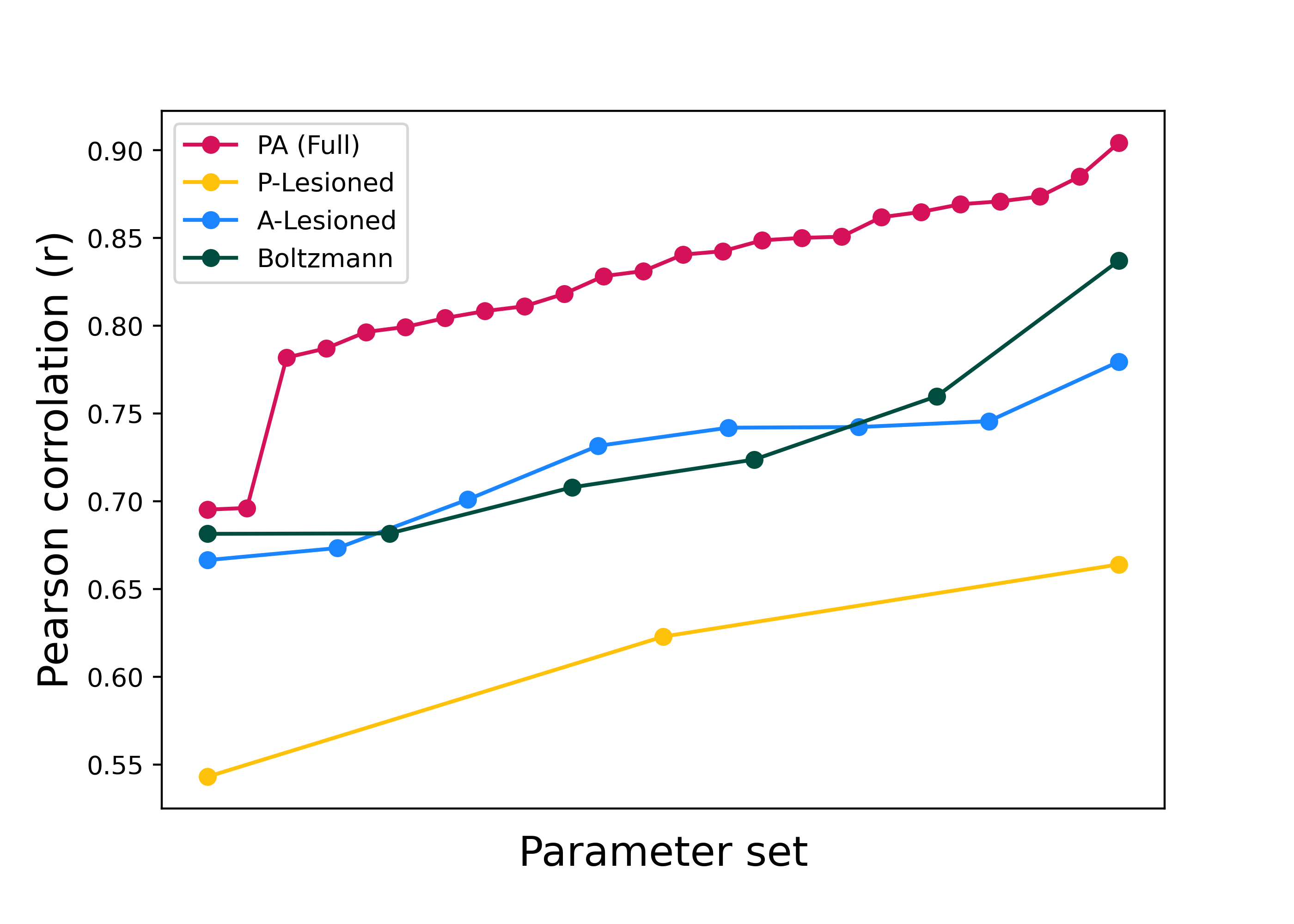}
         \caption{Sensitivity plot for Doors, Keys \& Gems.}
         \label{fig:sensitivity_dkg}
     \end{subfigure}
     \hspace{1pt}
     \begin{subfigure}[b]{0.4\textwidth}
         \centering
         \includegraphics[width=\textwidth]{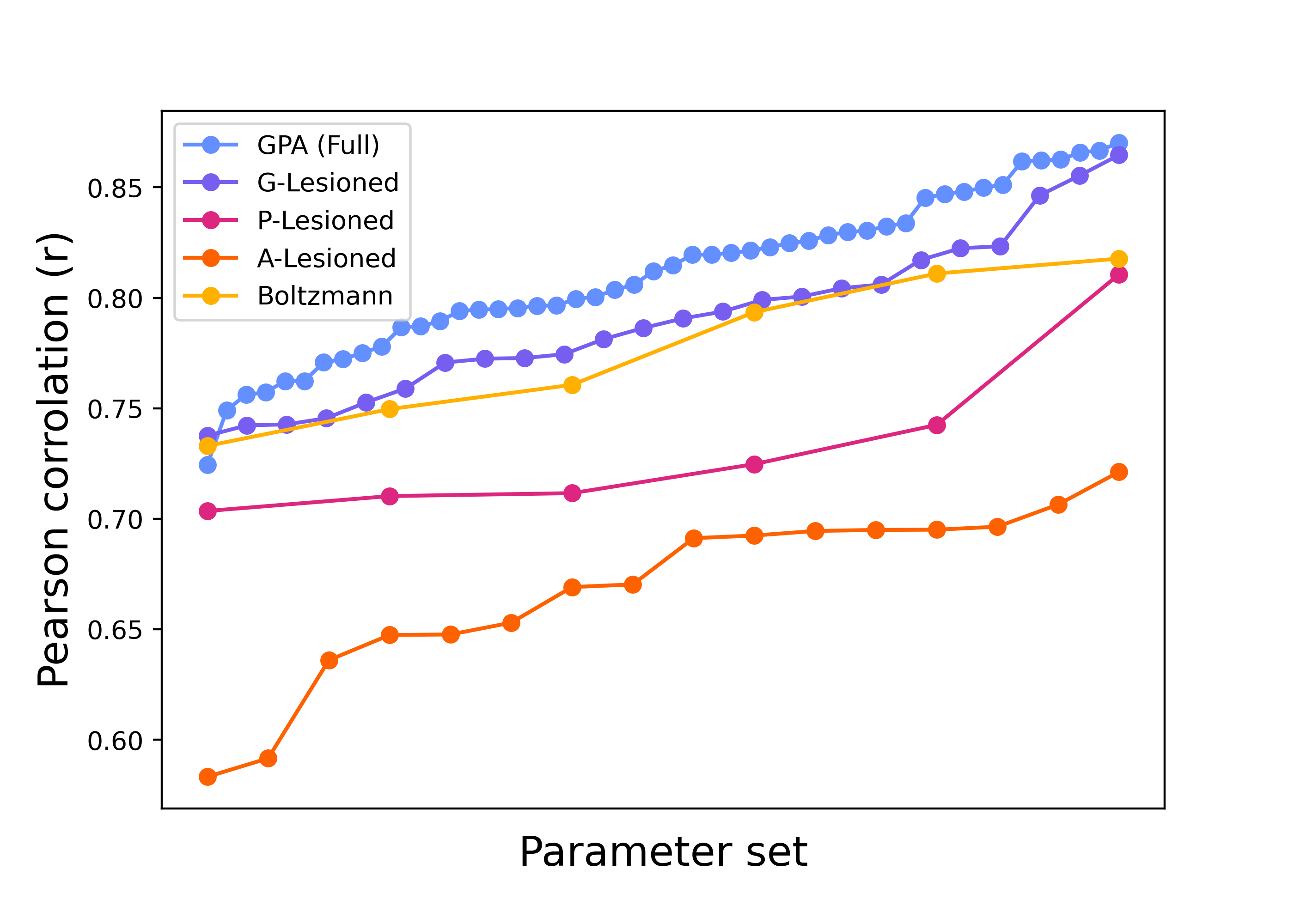}
         \caption{Sensitivity plot for Block Words.}
         \label{fig:sensitivity_bw}
     \end{subfigure}
     \caption{Sensitivity plots for human-model correlations (Pearson's $r$) with respect to model parameters. For each model, we plot $r$ values for each parameter set that we performed grid search over, ordering them from low to high in terms of correlation.}
    \label{fig:sensitivity}
\end{figure*}

\textbf{Results.}  The results of our analysis are presented in Figure \ref{fig:block-words-results}. We show an illustrative stimulus with goal confusion in Figure \ref{fig:block-words-results}(a). The panels show the agent intending to spell the word ``pear'' but misspells it as ``paer'' instead. Below these panels, we show average human inferences over time (with standard error ribbons), alongside the results for our bounded agent model and the baseline Boltzmann agent model. Humans are able to identify the true goal as soon as all the letters in ``pear'' are stacked, despite the wrong order ($t$=$7$). When the agent corrects their mistake  ($t$=$15$), humans remain confident in this inference. Our model, which encompasses goal, plan, and action (GPA) mistakes,  exhibits similar behaviour, assigning higher credence to ``pear'' from $t$=$7$ onwards. In contrast, the Boltzmann agent model fails to account for ``paer'' as a misspelling, assigning higher probability to a goal like ``reap'' because it takes less steps to reach from ``paer'' compared to ``pear''.

We compare average human vs. model inferences in Figure \ref{fig:block-words-results}(b). The left column shows the overall correlation across all 16 stimuli, indicating that our full model (GPA) best explains human inferences ($r=0.86 \pm 0.02$). The next few columns show correlations for each category of mistake. We see that our full model fit human data best when goal mistakes are present, compared to models which do not account for goal errors (G-lesioned, Boltzmann, etc.). The full model also did well with plan and action mistakes, though slightly out-performed by other models. Surprisingly, the Boltzmann model correlated highly even when trajectories exhibited short-sighted planning mistakes. We hypothesize that this is because the plan mistakes in our dataset are sufficiently short-lived (i.e. require backtracking only 1--2 actions) that they are readily attributed to action noise. In contrast, the model without action mistakes (A-lesioned) performed worst across the board, demonstrating the importance of modeling execution errors in this domain.

Sensitivity analysis indicated that the performance of our full model was robust across parameters. As Figure \ref{fig:sensitivity}(b) shows, the full model outperforms the P-Lesioned and A-Lesioned models across all parameter sets. For all but one parameter set, our full model also dominates the G-Lesioned and Boltzmann models in an ordinal sense: For each $k$, the top $k$th parameter set outperforms the corresponding parameter sets for the other two models.


\section{Discussion}

Our experimental results suggest that humans are robust to observed mistakes when inferring the goals of others. By comparing collected human inferences against inferences produced by our models, we find considerable support for our hypothesis that human inferences are better explained by a Bayesian Theory of Mind that accounts for mistaken goals, plans, and actions. In the Doors, Keys \& Gems domain, we find that models which only account for mistaken actions are not robust to mistaken plans, and vice versa, while in the Block Words domain, we find that models which do not account for goal confusion result in poor inferences when mistaken goals are observed. Together, these findings indicate that it is important to model \emph{distinct} errors at \emph{multiple} levels of cognition and action in order to understand others' goals, and that people intuitively distinguish these possibilities when observing sub-optimal behavior.

Nonetheless, many open questions remain. For one, what kinds of mistakes do humans understand beyond the three explored here? Prior research has also modeled biases like time inconsistency \cite{evans2016learning}, capability limitations \cite{lee2020getting}, and errors in environment models \cite{reddy2018you} when inferring the goals or preferences of others. Other cognitive limitations abound. Which of these do humans intuitively apprehend, and which do we not? For another, given the complexity of these models, how do human observers rapidly and tractably infer the goals and mistakes of others? Previous work has suggested that by assuming others are boundedly rational, the inference problem is made easier for observers, because they can perform online approximate inference over goals, plans and mistakes instead of exhaustively considering all possibilities in advance \cite{zhi2020online}. How well do these approximate inference algorithms explain the inferences made by individual humans? By answering these questions, we hope to build an even richer (theory) theory of our boundedly rational minds.

\section{Acknowledgements}

This work was funded in part by the DARPA Machine Common Sense program (030523-00001), DARPA SAIL-ON (HR001119S0038), Army Research Office MURI Grant No. W911NF-19-1-0057; the Aphorism Foundation, the Siegel Family Foundation, the MIT-IBM Watson AI Lab, and the Intel Probabilistic Computing Center.

\newpage

\renewcommand\thefigure{A\arabic{figure}}
\setcounter{figure}{0}

\section{Appendix}

\subsection{Experiment interface}

\begin{figure}[h]
    \centering
    \includegraphics[width=\columnwidth]{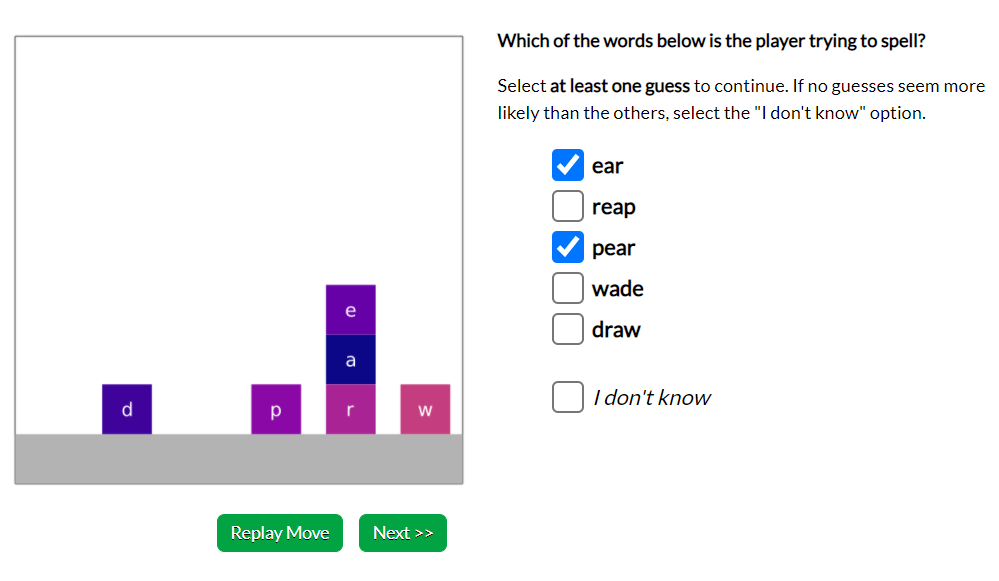}
    \caption{Experiment interface for Block Words.}
    \label{fig:interface}
\end{figure}

Figure \ref{fig:interface} shows the web interface used by participants. The interface displays animated stimuli in random order, each divided into segments. After observing each segment, participants provide goal inferences by selecting the goal(s) they believe to be the agent's most likely goal, then proceed to next segment. Responses are converted to uniform probability distributions over the model parameters (e.g., if two goals were selected, 50\% probability is allocated to each goal). If no single goal seems more likely than the others, participants can choose the \textit{I don't know} option, which is treated as a uniform distribution over all of the potential goals.

\subsection{Model fitting}

We fit our full model and the lesioned models to maximize human-model correlation by performing grid search over the following sets of parameters:
\begin{itemize}[itemsep=-0.5pt]
    \item Goal noise: $\epsilon_g \in [0.1, 0.2]$
    \item Parameters for the negative binomial  distribution over the planning budget $\eta$: $r \in [2, 4]$ and $q \in [0.9, 0.95]$.
    \item A$^*$ search noise: $\gamma \in [0.02, 0.5]$
    \item Action noise: $\epsilon_a \in [0.05, 0.1, 0.2]$
\end{itemize}
Across these models, we fixed the observation noise parameters to $\epsilon_o=0.05$ and $\sigma_o=0.25$ for Doors, Keys \& Gems, and $\epsilon_o=0.1$ for Block Words. For the A$^*$ search heuristic in Doors, Keys \& Gems, we used an optimistic maze distance metric which ignored the presence of locked doors, and the Fast Forward relaxed distance heuristic in Block Words \cite{hoffmann2001ff}. For the Boltzmann agent model, we fitted the inverse temperature parameter $\alpha$ over the range $[0.125, 0.25, 0.5, 1, 2, 4]$. Table \ref{tab:parameters} shows the fitted parameters.

To compute 95\% confidence intervals for the fitted $r$ values, we used the bootstrap method, resampling the human data without replacement 500 times and computing $r$ between the average human inferences and the model inferences. This produces an approximately normal histogram over $r$ values, for which we can compute a confidence interval which spans $95\%$ of the samples.

\begin{table}[t]
\centering
\begin{subtable}{\columnwidth}
\centering
\begin{tabular}{@{}lllllll@{}}
\toprule
 & $\epsilon_g$ & $r$ & $q$ & $\gamma$ & $\epsilon_a$ & $\alpha$ \\ \midrule
Full Model & -- & 2.0 & 0.9 & 0.5 & 0.05 & -- \\
P-Lesioned & -- & -- & -- & -- & 0.05 & -- \\
A-Lesioned & -- & 2.0 & 0.9 & 0.5 & -- & -- \\
Boltzmann & -- & -- & -- & -- & -- & 0.125 \\ \bottomrule
\end{tabular}
\caption{Model parameters for Doors, Keys \& Gems.}
\end{subtable}

\vspace{5pt}
\begin{subtable}{\columnwidth}
\centering
\begin{tabular}{@{}lllllll@{}}
\toprule
 & $\epsilon_g$ & $r$ & $q$ & $\gamma$ & $\epsilon_a$ & $\alpha$ \\ \midrule
Full Model & 0.2 & 2.0 & 0.9 & 0.02 & 0.05 & -- \\
G-Lesioned & -- & 2.0 & 0.9 & 0.50 & 0.20 &  \\
P-Lesioned & 0.2 & -- & -- & -- & 0.20 & -- \\
A-Lesioned & 0.2 & 4.0 & 0.9 & 0.02 & -- & -- \\
Boltzmann & -- & -- & -- & -- & -- & 2.0 \\ \bottomrule
\end{tabular}
\caption{Model parameters for Block Words.}
\end{subtable}

\caption{Best-fitting model parameters across domains and model types: goal noise $\epsilon_g$, planning budget parameters $r$ and $q$, search noise $\gamma$, action noise $\epsilon_a$, and inverse temperature for the Boltzmann model $\alpha$.}
\label{tab:parameters}
\end{table}

\subsection{Additional qualitative results}

In addition to collecting human goal inferences, we also asked participants to identify whether they believed a mistake occurred after viewing each of the last two stimuli they were presented with. If they stated that a mistake occurred, they were asked to provide a qualitative description of the mistake. For the Doors, Keys \& Gems domain, we found that 100\% of participants responded correctly when mistakes were absent, and that 81\% of participants detected a mistake when present. For Block Words domain, 100\% of participants responded correctly when mistakes were absent, and 70\% detected a mistake when present, but sometimes could not remember details well enough to describe the mistake. In addition, participants accurately identified action mistakes only 20\% of the time.  These results validate the intuitive hypothesis that humans can recognize mistakes when they occur, but also suggest that conscious recognition of mistakes is not necessary for accurate goal inference.

\end{document}